\documentclass[journal,comsoc]{IEEEtran}

\usepackage[T1]{fontenc}

\ifCLASSINFOpdf
\else
\fi

\usepackage{amsmath}
\usepackage{verbatim}
\usepackage{bm}
\usepackage{graphicx}
\usepackage{enumerate}
\usepackage{amsfonts}
\usepackage{cite}
\usepackage{color}
\usepackage{amsbsy}
\usepackage{amsmath}
\usepackage{indentfirst}
\usepackage{booktabs}

\usepackage[cmintegrals]{newtxmath}

\usepackage[utf8]{inputenc} 
\usepackage[T1]{fontenc}    
\usepackage{url}            
\usepackage{nicefrac}       
\usepackage{microtype}      
\usepackage[dvipsnames]{xcolor}         
\usepackage{interval}
\usepackage{bm}
\usepackage{bbm}
\usepackage{mathtools}
\usepackage{footmisc}
\usepackage{subfig}
\usepackage{xspace}
\usepackage[inline]{enumitem}
\usepackage[ruled,vlined]{algorithm2e}
\usepackage{soul}

\newcommand{\system}{CDCL\@\xspace}

\newcommand*{\eg}{\emph{e.g.}\@\xspace}
\newcommand*{\ie}{\emph{i.e.}\@\xspace}
\newcommand*{\etc}{\emph{e.t.c.}\@\xspace}

\newcommand{\zz}{{\bm z}}

\DeclareMathOperator{\E}{\mathbb{E}}
\DeclareMathOperator*{\minimize}{minimize}

\begin{document}

\title{Cross-domain Contrastive Learning for\\ Unsupervised Domain Adaptation}

\author{Rui~Wang,
        Zuxuan~Wu,
        Zejia~Weng,
        Jingjing~Chen,
        Guo-Jun~Qi,~\IEEEmembership{Fellow,~IEEE,}
        and~Yu-Gang~Jiang
\thanks{Rui Wang, Zuxuan Wu, Zejia Weng, Jingjing Chen and Yu-Gang Jiang are with Shanghai Key Lab of Intelligent Information Processing, School of Computer Science, Fudan University and Shanghai Collaborative Innovation Center on Intelligent Visual Computing.}
\thanks{Guo-Jun~Qi is with the Seattle Cloud Lab, Futurewei Technologies, Bellevue, WA 98004.}
\thanks{Corresponding author: Zuxuan Wu.}
}

\markboth{IEEE TRANSACTIONS ON MULTIMEDIA}%
{Shell \MakeLowercase{\textit{et al.}}: Bare Demo of IEEEtran.cls for IEEE Journals}

\maketitle

\begin{abstract}
Unsupervised domain adaptation (UDA) aims to transfer knowledge learned from a fully-labeled source domain to a different unlabeled target domain. Most existing UDA methods learn domain-invariant feature representations by minimizing feature distances across domains. In this work, we build upon contrastive self-supervised learning to align features so as to reduce the domain discrepancy between training and testing sets. Exploring the same set of categories shared by both domains, we introduce a simple yet effective framework CDCL, for domain alignment. In particular, given an anchor image from one domain, we minimize its distances to cross-domain samples from the same class relative to those from different categories. Since target labels are unavailable, we use a clustering-based approach with carefully initialized centers to produce pseudo labels. In addition, we demonstrate that CDCL is a general framework and can be adapted to the data-free setting, where the source data are unavailable during training, with minimal modification. We conduct experiments on two widely used domain adaptation benchmarks, i.e., Office-31 and VisDA-2017, for image classification tasks, and demonstrate that CDCL achieves state-of-the-art performance on both datasets.
\end{abstract}


\IEEEpeerreviewmaketitle

\section{Introduction}

At the heart of many machine learning and computer vision tasks is to learn robust feature representations that generalize well to novel testing samples. However, state-of-the-art deep learning models still suffer from significant performance drops even when the testing distribution slightly drifts from the training distribution. To mitigate this issue, unsupervised domain adaptation \cite{ben2010theory,ben2007analysis,bousmalis2017unsupervised,long2015learning,ganin2015unsupervised,saenko2010adapting} aims to reduce the discrepancy between training and testing, which is also known as domain shifts. This is generally achieved by aligning the distribution of a labeled training set (source domain) with that of an unlabeled testing set (target domain) \cite{long2015learning,long2017deep}. In particular, feature alignment aims to minimize carefully designed metrics like Maximum Mean Discrepancies (MMD) \cite{gretton2006kernel}, covariances \cite{sun2016return,sun2016deep}, and adversarial loss functions \cite{ganin2015unsupervised,tzeng2017adversarial} such that the distances between training and testing distributions are reduced.

The idea of reducing feature distance in UDA tasks is similar in spirit to recent advances in self-supervised contrastive learning, which pulls an image to be closer to its own augmented copy on a hypersphere compared to other images. In this paper, we ask the following question: can we leverage contrastive learning that produces decent feature representations in a variety of downstream tasks \cite{he2020momentum,chen2020simple,chen2020improved} for domain alignment in unsupervised domain adaptation? While appealing, this is non-trivial as in standard contrastive learning a positive pair can be naturally generated considering two related views of the same image, since they contain the same content but are transformed with different augmentations. In domain adaptation, it is not clear how to form positive and negative pairs in order to align feature distributions.

Exploring the fact that categories are shared between the source and target domain, we propose to align feature representations conditioned on class information to learn domain-invariant features. In particular, we argue that samples within the same class should be closer to each other while samples from different categories should lie far apart, even when they are from different domains. 

\begin{figure*}[t] \centering
  \resizebox{0.92\linewidth}{!}{\includegraphics[width=0.92\linewidth]{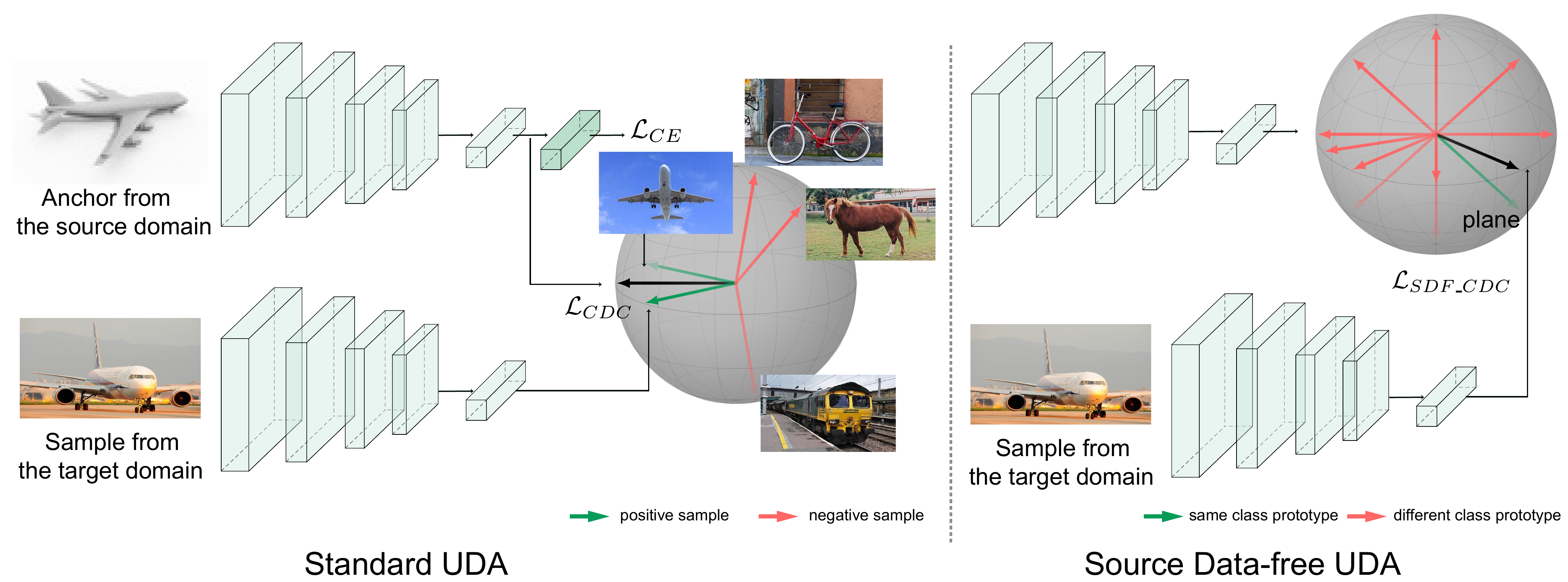}}
  \caption{A conceptual overview of our approach in the standard UDA setting (left) and the data-free setting (right). Left: Given an anchor image from the source domain, we pull its feature to be closer to samples of the same category from the target domain while pushing apart its feature with those from a different class in the target domain. Right: when source data are not available, we replace source samples with prototypical features derived from a pre-trained model in the source domain. See texts for more details.}
  \label{fig:approach}
\end{figure*}

In light of this, we introduce \system, a simple yet effective framework for unsupervised domain adaptation under both standard and data-free settings. As shown in Figure~\ref{fig:approach}, given an anchor image from the source domain, we randomly select samples from the target domain that belong to the same class as the anchor to form positive pairs, based on pseudo labels of target samples in lieu of manual labels. We minimize the distance of all positive pairs relative to negative pairs, which are formed by cross-domain samples from different categories. Since labels are not available for the target domain, we generate pseudo labels with k-means clustering, whose initial clusters are set to class prototypes learned on the source domain. Through minimizing feature distance with the proposed cross-domain contrastive loss, \system produces domain-invariant features. We further show \system can be conveniently adapted to the newly proposed data-free scenario \cite{liang2020we}, where the source data are not available, by replacing sample features with prototypical features.

We conduct extensive experiments on two widely used domain adaptation benchmarks, \ie, Office-31 \cite{saenko2010adapting} and VisDA \cite{peng2017visda} and demonstrate that our method achieves state-of-the-art performance on both datasets. We further show that \system can effectively produce domain-invariant features even when source data are not available. We also conduct a set of ablation experiments to validate the effectiveness of different components of our approach.

\section{Related Work}

\textbf{Unsupervised domain adaptation (UDA).} Existing UDA methods focus on learning domain-invariant feature representations. One direction of UDA methods is minimizing the discrepancy between different domains  \cite{long2015learning,long2017deep,sun2016return,sun2016deep,zellinger2017central}. In early approaches \cite{long2015learning,long2017deep}, MK-MMD and joint MMD are employed to measure the discrepancy between the source domain and target domain. Besides, higher-order statistics and other well-designed discrepancies are utilized in \cite{zellinger2017central,lee2019sliced,zhang2019bridging,li2021discriminative,yan2019weighted}. To leverage category information for domain alignment, CAN \cite{kang2019contrastive} introduces intra- and inter-class domain discrepancies. In \cite{kang2019contrastive}, class information from the target domain is obtained by K-means clustering, which is similar to the generation of pseudo labels in our method. However, CAN is unpractical for source data-free UDA due to the utilization of source data when minimizing the domain discrepancy between source samples and target samples. Another direction is to design an adversarial optimization objective for a domain discriminator and to obtain domain-invariant representations by adversarial learning \cite{ganin2015unsupervised,long2018conditional}. GVB-GD \cite{cui2020gradually} promotes adversarial domain adaptation with a gradually vanishing bridge mechanism. GSDA \cite{hu2020unsupervised} implements hierarchical domain alignment with multiple adversarial discriminators. Recently, in addition to discrepancy-based methods and adversarial-based methods, there are other UDA methods, e.g., domain-adaptive dictionary learning \cite{zheng2019multi}, multi-modality representation learning \cite{ma2019deep} and feature disentanglement \cite{jin2021style}. Some recent approaches \cite{xu2019larger,chang2019domain,cui2020towards} also explore feature norm or batch norm for UDA. SAFN \cite{xu2019larger} enlarges feature norms of different domains to improve the transferability of features. \cite{chang2019domain} uses the domain-specific batch normalization and \cite{cui2020towards} performs batch nuclear-norm maximization to generate discriminative and diverse predictions. To improve feature discriminability, BSP \cite{chen2019transferability} penalizes the largest singular values and ADR \cite{saito2018adversarial} applies dropout on the classifier. To avoid ambiguous target features, MCD \cite{saito2018maximum} maximizes the discrepancy between two classifiers and STAR \cite{lu2020stochastic} samples classifiers from Gaussian distributions without more parameters. In this paper, we align features with contrastive learning, which is simple and easy to optimize. 

\textbf{Contrastive learning.} Great progress in unsupervised representation learning has been achieved by self-supervised contrastive learning \cite{oord2018representation,chen2020simple,he2020momentum,chen2020improved}. The standard approach of contrastive learning is to learn discriminative representations by pulling together positive pairs and pushing apart negative pairs. In self-supervised learning methods  \cite{chen2020simple,he2020momentum,chen2020improved}, the positive pairs are produced by creating different augmented views of each sample, while negative pairs can be randomly chosen from different samples. Instance discriminative representations learned by self-supervised contrastive learning can be transferred well to downstream tasks with fine-tuning. However, without task-specific semantic information, representations with intra-class compactness and inter-class discrimination can not be learned through instance-level contrastive learning. Recently, supervised contrastive learning \cite{khosla2020supervised} leverages category labels to compose positive and negative pairs and achieves promising performance on fully-supervised image classification.  \cite{ge2020self} proposes a self-paced contrastive learning framework for domain adaptive object re-ID with multi-level supervision in each domain. Nonetheless, feature alignment, which is critical for domain adaptation methods, is not considered in these contrastive learning methods. There are some approaches \cite{thota2021contrastive,dai2020contrastively,toldo2021unsupervised} applying contrastive learning for other UDA tasks. \cite{thota2021contrastive} simply performs contrastive learning on each domain independently and minimizes MMD to reduce the domain gap. Nevertheless, neither domain alignment nor class alignment is considered in the contrastive loss of \cite{thota2021contrastive}. CoSCA \cite{dai2020contrastively} enhances the MCD \cite{saito2018maximum} framework with contrastive loss that separates the ambiguous target samples, and uses MMD to obtain better global domain alignment. As mentioned in \cite{saito2018maximum}, the classification loss on the source data is necessary for MCD when maximizing the discrepancy of two classifiers on the target data. Therefore, CoSCA can not be directly transferred to source data-free UDA. \cite{toldo2021unsupervised} proposes an effective feature clustering-based strategy to capture the different semantic modes of the feature distribution and group features of the same class into tight and well-separated clusters for improved unsupervised domain adaptation in semantic segmentation. However, it confuses the data of the source and target domains during the clustering-based training process, making it impossible to fit in source data-free setting. It is worth noting a concurrent work explores the idea of contrastive learning for domain adaptation \cite{park2020joint}. However, the approach mixes samples from all domains and thus can only be used in the standard UDA setting. Besides, the results of our ablation study show that if ignoring the domain-level information and simply considering label information in the contrastive loss like \cite{park2020joint}, the performance will degrade.

\textbf{Source Data-free UDA.} Recently, due to the concern of source data privacy in the realistic applications of UDA methods, source data-free UDA has been proposed by  \cite{liang2020we}. The main challenge of source data-free UDA is that a pre-trained model on the source domain should be adapted to the target domain without access to source data. Based on hypothesis transfer learning, \cite{liang2020we} proposes a self-training framework with mutual information maximization and pseudo-labeling strategy. In  \cite{li2020model}, a collaborative class conditional generative adversarial network is employed to avoid the usage of source data with target data generation and model adaptation. In this paper, instead of using entropy minimization, we leverage contrastive learning for cross-domain alignment.

\section{Methodology}

Unsupervised domain adaptation aims to transfer models learned on a labeled source domain to an unlabeled target domain, whose data distribution is different from that of the source domain. During training, UDA assumes access to all labeled samples in the source domain as well as unlabeled images from the target domain. Formally, given a fully-labeled source domain dataset with $N_s$ image and label pairs $D_s = (\mathcal X_s,\mathcal Y_s)= \{(x_s^i,y_s^i)\}^{N_s}_{i=1}$, and an unlabeled dataset in a target domain  with $N_t$ images $D_t = \mathcal X_t= \{x_t^i\}^{N_t}_{i=1}$, both $\{x_s^i\}$ and $\{x_t^i\}$ belong to the same set of $M$ predefined categories. We use $y_s^i \in \{ 0, 1, \ldots, M-1 \}$  to represent the label of the $i$-th source sample while the labels of target samples are unknown during training. UDA aims to predict labels of testing samples in the target domain using a model $f_t: \mathcal X_t \to \mathcal Y_t$ trained on $D_s \cup D_t$.  The model, parameterized by ${\bm \theta}$ consists of a feature encoder $g: \mathcal X_t \to \mathbb{R}^d$ and a classifier $h: \mathbb{R}^d \to \mathbb{R}^M$, where $d$ is the dimension of features produced by the encoder.

Our goal is to align feature distributions between the source and the target domain through contrastive self-supervised learning. To this end, we first briefly review contrastive learning, and then introduce \system that forms positive and negative pairs in a cross-domain manner to learn domain-invariant features. Finally, we show that the proposed approach is not only suitable for standard UDA but can also be applied to data-free scenarios, where the source data are unavailable during training.

\subsection{Contrastive Learning with InfoNCE}
State-of-the-art contrastive learning frameworks typically use the N-pair loss \cite{sohn2016improved}, also known as the InfoNCE \cite{oord2018representation,he2020momentum} and NT-Xent loss \cite{chen2020simple}, to minimize the distance of a positive pair relative to all other pairs. More formally, let $\bm u$ and $\bm v$ denote $\ell_2$-normalized feature representations of a pair and the loss function is then defined as:

\begin{align}
\mathcal{L}=- \sum\limits_{\bm v^+ \in V^+} \log \frac{\exp (\bm u^\top \bm v^+ / \tau)}{\exp (\bm u^\top \bm v^+ / \tau) + \sum_{\bm v^-\in V^-}\exp(\bm u^\top \bm v^- /\tau)},
\label{eqn:npair}
\end{align}

where $\bm v^+ \in V^+$ and $\bm v^- \in V^-$ represent the positive and negative samples with respect to $\bm u$, and $\tau$ is a temperature parameter that is manually set. In practice, a positive pair is derived by two random data augmentations (\ie, blurring and color jittering, \etc) operated on the same image sample, resulting in two correlated views. In contrast, in domain adaptation, it remains unclear how to obtain positive and negative pairs for feature alignment.

\subsection{Cross-domain Contrastive Learning}
We now introduce how to form pairs to learn domain-invariant features with contrastive learning. Since samples from the source domain and the target domain belong to the same set of classes in current UDA settings, we build upon this assumption to reduce domain shift. More specifically, we hypothesize that samples within the same category are close to each other while samples from different classes lie far apart, regardless of which domain they come from.  More formally, we consider the $\ell_2$-normalized features ${\bm z}^i_t$ from the $i$-th sample ${\bm x}_t^i$ in the target domain as an anchor, and it forms a positive pair with a sample in the same class from the source domain, whose features are denoted as  ${\bm z}_s^p$, we formulate the cross-domain contrastive loss as:

\begin{equation}
  \mathcal{L}_{CDC}^{t, i} = - \frac{1}{|P_s(\hat{y}^i_t)|} \sum\limits_{p \in P_s(\hat{y}^i_t)} \log \frac{\exp ({\zz_t^i}^\top \zz_s^p / \tau) }{\sum\limits_{j \in I_s} \exp ({\zz_t^i}^\top \zz_s^j / \tau)}
  \label{eqn:cdc}
\end{equation}

where $I_s$ denotes the set of source samples in a mini-batch and $P_s(\hat{y}^i_t) = \{k \ | \ y_s^k = \hat{y}^i_t \}$ indicates the set of positive samples from the source domain that share the same label with the target anchor $x_t^i$. Since we do not have access to labels of target samples, we use estimated pseudo labels $\hat{y}^i_t$ (as will be introduced below) to generate pairs. The cross-domain loss forces intra-class distance to be smaller than inter-class distance for samples from different domains so as to reduce domain shift. It is worth pointing out that compared to the standard InfoNCE loss, we sum over all samples in a mini-batch from the source domain that belong to the same category as the anchor $x_t^i$, which could reduce sampling variance. 

In Eqn.~\ref{eqn:cdc}, we consider samples from the target domain as anchors. Alternatively, we can use source samples as anchors and compute $\mathcal{L}_{CDC}^{s, i}$ similarly  by setting $P_s(y^i_s) = \{k \ | \ \hat{y}_t^k = y^i_s \}$. Then, we combine $\mathcal{L}_{CDC}^{s, i}$ with $\mathcal{L}_{CDC}^{t, i}$ to derive the cross-domain contrastive loss as follows:

\begin{equation}
\label{eqn:cdc_bi}
  \mathcal{L}_{CDC} = \sum\limits_{i=1}^{N_s} {\mathcal{L}_{CDC}^{s, i}} + \sum\limits_{i=1}^{N_t} {\mathcal{L}_{CDC}^{t, i}}.
\end{equation}

The cross-domain contrastive loss aligns features in a bi-directional manner by using anchors from both domains for improved performance. Finally, combining the cross-domain contrastive loss with a standard cross-entropy loss $\mathcal{L}_{CE}$ enforced on the source domain, we have the final objective function for training:
\begin{align}
  \minimize_{\bm \theta} \, \mathcal{L}_{CE}({\bm \theta};D_s) + \lambda \mathcal{L}_{CDC} ({\bm \theta};D_s, D_t),
  \label{eqn:uda}
\end{align}
where $\lambda$ controls the trade-off between the two loss terms and ${\bm \theta}$ denotes the parameters to be optimized.

\SetKwInput{KwInput}{Input}                
\SetKwInput{KwOutput}{Output}              

\begin{algorithm}
\SetAlgoLined
\KwResult{$\bm \theta$ for the prediction model $f$}
\KwInput{unlabeled target dataset $D_t = \mathcal X_t$, source dataset $D_s = (\mathcal X_s,\mathcal Y_s)$, model $f=h \circ g$, max epoch $E$, iterations per epoch $K$}
    Initialize encoder $g$ with ImageNet pre-trained weights  \\
\For{$e = 1$ to $E$}{
   Initialize cluster centers with source class prototypes using Eqn.~\ref{eqn:cluster_uda_sdf} \\
    Perform K-means clustering on target data $\mathcal X_t$, obtain pseudo labels $\hat{y}_t^i$ \\
    \For{$k = 1$ to $K$}  
    { 
           Sample batch $(x_s^i, y_s^i)$ from $D_s$ and compute $\mathcal{L}_{CE}$ \\
           Sample batches $(x_s^j, y_s^j)$ and $(x_t^j, \hat{y}_t^j)$ from $D_s$ and $D_t$ \\
           Compute $\mathcal{L}_{CDC}$ using Eqn.~\ref{eqn:cdc_bi} \\
           Back-propagate and update model $f$ via Eqn.~\ref{eqn:uda}
    }
}
\caption{Pseudo code of \system for standard UDA.}
\label{alg:alg1}
\end{algorithm}

\begin{algorithm}
\SetAlgoLined
\KwResult{$\bm \theta$ for the prediction model $f$}
\KwInput{unlabeled target dataset $D_t = \mathcal X_t$, model $f=h \circ g$ pre-trained on source dataset $D_s$, max epoch $E$, iterations per epoch $K$}
    Freeze the parameters of classifier $h$ \\
\For{$e = 1$ to $E$}{
        Initialize cluster centers with source class prototypes using Eqn.~\ref{eqn:cluster_uda} \\
    Perform K-means clustering on target data $\mathcal X_t$, obtain pseudo labels $\hat{y}_t^i$ \\
    \For{$k = 1$ to $K$}  
    { 
            Sample batch $(x_t^i, \hat{y}_t^i)$ from $D_t$ and compute $\mathcal{L}_{SDF-CDC}^{t, i}$ using Eqn.~\ref{eqn:cdc_sdf} \\
            Back-propagate and update model $f$ via Eqn.~\ref{eqn:uda_sdf_obj}
    }
}
\caption{Pseudo code of \system for source data-free UDA. }
\label{alg:alg2}
\end{algorithm}

\subsection{Pseudo Labels for the Target Domain }
 Ground-truth labels from the target domain are not available during training, and thus we leverage k-means clustering to produce pseudo labels \cite{kang2019contrastive, liang2020we}, forming pairs for cross-domain contrastive learning. Since K-means is sensitive to initialization, using randomly generated clusters fails to guarantee related semantics with respect to predefined categories. To mitigate this issue, we set the number of clusters to the number of classes $M$ and use class prototypes from the source domain as initial clusters. The benefits of initializing the cluster centers with class prototypes are twofold: \romannumeral1) source class prototypes can be seen as the approximation of target class prototypes, since features used are high-level and contain semantics information (\romannumeral2) with the alignment of samples in the same category by CDCL, this approximation will be more accurate as the training continues. More formally, we first compute the centroid of source samples in each category as the corresponding class prototype and the initial cluster center ${\bm O}_t^m$ for the $m$-th class is defined as:
\begin{equation}
  O_t^m \gets O_s^m = \E_{i \sim D_s,y_s^i = m} \zz_s^i.
  \label{eqn:cluster_uda}
\end{equation}

Given features from the target domain, we then perform spherical K-means clustering using these carefully initialized centers. When determining the assignment of each target sample, cosine similarity is adopted to measure the distance between the target feature $\zz_t^i$ and the $m$-th cluster center ${\bm O}_t^m$. Once clustering is finished, each sample in the target domain ${\bm x}_t^i$ is associated with a pseudo label $\hat{y}_t^i$. To reduce the noise in target pseudo labels, we remove the ambiguous samples far from its assigned clustered centers. Concretely, one target sample will be removed when the cosine similarity between its feature and its assigned cluster center is below a manually set threshold $d$.

\subsection{Source Data-free UDA}
In this section, we demonstrate that \system can be easily adapted to a newly introduced source data-free setting \cite{liang2020we}, where a model trained on the source domain is provided yet source data are unavailable due to corruption or privacy concerns. Formally, the goal is to learn a model $f_t:X_t \to Y_t$ and predict $\{ y_t^i \}_{i=1}^{N_t}$ with only unlabeled target data $D_t$ and a pre-trained source model $f_s: X_s \to Y_s$. The pre-trained source model is obtained by minimizing the cross-entropy loss on the source samples.

It is difficult for most previous UDA methods to adapt to source data-free UDA. For discrepancy-based methods, the predefined domain discrepancies are statistics that should be measured between source samples and target samples. Similarly, for adversarial-based methods, the adversarial discriminators need to be trained with source samples and target samples. Without access to source data, both strategies are not applicable to perform adaptation to the target domain. Besides, for the standard UDA setting, many UDA methods assume that the same feature encoder is shared on the source and target domains. However, this constraint may be hard to implement under source data-free setting since the feature encoder can not be trained on the source and target domain simultaneously. Some UDA methods, \eg, DSBN \cite{chang2019domain}, prove that a domain-specific module in the feature encoder can improve the performance of domain adaptation. Therefore, it is practical to remove the parameter-sharing constraint for feature encoders under source data-free setting.

For CDCL, the lack of samples from the source domain $D_s$ makes it challenging to (1) form positive and negative pairs and (2) to compute source class prototypes. We address this issue by replacing source samples with classifier weights from the trained model $f_s$.  The intuition is that the weight vectors in the classifier layer of a pre-trained model can be regarded as prototypical features of each class learned on the source domain. In particular, we first remove the bias of the fully-connected layer and perform normalization for the classifier. 

We use ${\bm w}_s^m$ to denote the weight vector of the $m$-th class in the  classification layer ${\bm W}_s = [{\bm w}_s^1, \ldots, {\bm w}_s^M]$ learned on the source domain. Since the weights are normalized, we use them as class prototypes. When adapting to the target domain, we freeze the parameters of the classifier layer to keep the source class prototypes and only train the feature encoder. Through replacing the source samples with source class prototypes, the cross-domain contrastive loss under the source data-free setting can be written as:

\begin{equation}
  \mathcal{L}_{SDF-CDC}^{t, i} = - \sum\limits_{m=1}^{M} \mathbf{1}_{\hat{y}^i_t = m} \log \frac{\exp ({\zz_t^i}^\top {\bm w}_s^m / \tau) }{\sum\limits_{j=1}^{M} \exp ({\zz_t^i}^\top {\bm w}_s^j / \tau)}.
  \label{eqn:cdc_sdf}
\end{equation}

Similarly, we estimate labels for samples in the target domain with clustering. However, it is not feasible to compute class prototypes using samples anymore. Instead, we replace Eqn.~\ref{eqn:cluster_uda} with class weights:

\begin{equation}
O_t^m \gets O_s^m = {\bm w_s}^m 
\label{eqn:cluster_uda_sdf}
\end{equation}

The final objective of source data-free UDA is:

\begin{equation}
 \minimize \, \sum\limits_{i=1}^{N_t} \mathcal{L}_{SDF-CDC}^{t, i}.
 \label{eqn:uda_sdf_obj}
\end{equation}
Compared to Eqn.~\ref{eqn:uda}, the cross-entropy loss is not used since the source data are no longer available for supervised training.

\section{Experiments}

\begin{table*}[h]
  \caption{Accuracy(\%) on VisDA-2017 for unsupervised domain adaptation (ResNet-101). $\dagger$ denotes that this method is developed under the source data-free UDA setting.}
  \label{tab:visda}
  \centering
  \setlength{\tabcolsep}{0pt} 
   \begin{tabular*}{\linewidth}{@{\extracolsep{\fill}\quad}lcccccccccccccc}
    \toprule
    Method &  \rotatebox{90}{plane} & \rotatebox{90}{bcycl} & \rotatebox{90}{bus} & \rotatebox{90}{car} &  \rotatebox{90}{horse} & \rotatebox{90}{knife} & \rotatebox{90}{mcycl} & \rotatebox{90}{person} & \rotatebox{90}{plant} & \rotatebox{90}{sktbrd} & \rotatebox{90}{train} & \rotatebox{90}{truck} & Avg \\
    \midrule
    ResNet-101 \cite{he2016deep}   & 55.1 & 53.3 &  61.9 &  59.1 &  80.6 & 17.9 &  79.7 & 31.2 &  81.0 & 26.5 & 73.5 & 8.5 & 52.4 \\
    \midrule
    DANN \cite{ganin2015unsupervised}      &  81.9 & 77.7 &  82.8 &  44.3 &  81.2 & 29.5 &   65.1 &  28.6 &  51.9 &  54.6 & 82.8 & 7.8 & 57.4 \\
    DAN \cite{long2015learning}      &  87.1 & 63.0 &  76.5 &  42.0 &  90.3 & 42.9 &   85.9 & 53.1 &  49.7 &  36.3 & 85.8 & 20.7 & 61.1 \\
    ADR \cite{saito2018adversarial}    &  87.8 &  79.5 &  83.7 &   65.3 &  92.3 & 61.8 &  88.9 &  73.2 &   87.8 &  60.0 &  85.5 & 32.3 & 74.8 \\
    CDAN \cite{long2018conditional}   &  85.2 &  66.9 &  83.0 &   50.8 &  84.2 & 74.9 &  88.1 & 74.5 &  83.4 &  76.0 & 81.9 &  38.0 & 73.7 \\
    CDAN+BSP \cite{chen2019transferability} &  92.4 & 61.0 &  81.0 &  57.5 &  89.0 & 80.6 & 90.1 & 77.0 &  84.2 & 77.9 & 82.1 & 38.4 & 75.9 \\
    SAFN \cite{xu2019larger}  & 93.6 & 61.3 &  84.1 &  70.6 &  94.1 & 79.0 &  \textbf{91.8} & 79.6 &  89.9 & 55.6 & 89.0 & 24.4 & 76.1 \\
    SWD \cite{lee2019sliced}  & 90.8 &  82.5 & 81.7 & 70.5 &  91.7 & 69.5 & 86.3 & 77.5 & 87.4 & 63.6 & 85.6 & 29.2 & 76.4 \\
    MSTN+DSBN \cite{chang2019domain} & 94.7 & 86.7 &  76.0 &  72.0 &  95.2 & 75.1 &  87.9 & 81.3 & 91.1 & 68.9 & 88.3 & 45.5 & 80.2 \\
    STAR \cite{lu2020stochastic}  & 95.0 & 84.0 &  84.6 &   73.0 &  91.6 & 91.8 &  85.9 &  78.4 &   94.4 & 84.7 & 87.0 &  42.2 & 82.7 \\
    CoSCA \cite{dai2020contrastively} & 95.7 & 87.4 &  85.7 & 73.5 &  95.3 & 72.8 &  91.5 &  84.8 &   94.6 & 87.9 & 87.9 &  36.8 & 82.9 \\
    CAN \cite{kang2019contrastive}     & 97.0 & 87.2 &  82.5 &  74.3 &  \textbf{97.8} & 96.2 &  90.8 &  80.7 &  \textbf{96.6} &  \textbf{96.3} & 87.5 & 59.9 &  87.2 \\
    JCL \cite{park2020joint} & 97.0 & \textbf{91.3} &  84.5 &  66.8 &  96.1 & 95.6 &  89.8 &  81.5 & 94.7 & 95.6 & 86.1 & \textbf{71.8} & 87.6 \\
    \midrule
    CDCL (ours)  & \textbf{97.4} & 89.5 & \textbf{85.9} & \textbf{78.2} & 96.4 & \textbf{96.8} & 91.4 & \textbf{83.7} & 96.3 & 96.2 & \textbf{89.7} & 61.6 & \textbf{88.6} \\
    \midrule
    SHOT  \cite{liang2020we} $\dagger$      & 94.3 & 88.5 &  80.1 &   57.3 &  93.1 & 94.9 &  80.7 & 80.3 &  91.5 & 89.1 & 86.3 & 58.2 & 82.9 \\
    ModelAdapt  \cite{li2020model} $\dagger$    &  94.8 & 73.4 &  68.8 &  \textbf{74.8} &  93.1 & 95.4 &  88.6 &  84.7 &  89.1 & 84.7 &  83.5 & 48.1 &  81.6 \\
    CDCL (ours) $\dagger$    & \textbf{97.3} & \textbf{90.5} &  \textbf{83.2} &  59.9 & \textbf{96.4} & \textbf{98.4} & \textbf{91.5} & \textbf{85.6} & \textbf{96.0} & \textbf{95.8} & \textbf{92.0} & \textbf{63.8} & \textbf{87.5} \\
    \bottomrule
  \end{tabular*}
\end{table*}

\subsection{Datasets and Compared Approaches}
\label{sec:data}
We use two public benchmarks to evaluate our method for unsupervised domain adaptation under both standard and data-free settings.

\textbf{VisDA-2017} \cite{peng2017visda} is a challenging large-scale benchmark including 12 classes from two domains:  the source domain with 152,397 synthetic images, and the target domain contains 55,388 real-world images. Our method is evaluated on the synthesis-to-real domain adaptation task. 

\textbf{Office-31} \cite{saenko2010adapting} is a common DA benchmark which contains 4,110 images from three distinct domains, \ie, Amazon (\textbf{A} with 2,817 images), DSLR (\textbf{D} with 498 images) and Webcam (\textbf{W} with 795 images). Each domain consists of 31 object categories. Our method is evaluated by performing domain adaptation on each pair of domains, which generates 6 different tasks.

\textbf{Compared Approaches.} We first report the results of a model trained on the source domain only, and compare with the following state-of-the-art approaches:\begin{enumerate*}[label=(\alph*)]
  \item DANN \cite{ganin2015unsupervised}, which utilizes a domain discriminator with adversarial optimization objective to reduce the domain gap. 
  \item DAN \cite{long2015learning} and JAN \cite{long2017deep},  which learn domain-invariant features by minimizing MK-MMD and Joint MMD.
  \item ADR \cite{saito2018adversarial}, which encourages the encoder to generate more discriminative features by using dropout on the classifier.
  \item SAFN \cite{xu2019larger}, which adapts the feature norms of different domains to a large range of values.
  \item SWD \cite{lee2019sliced}, which measures the dissimilarity between the output of classifiers with sliced Wasserstein discrepancy.
  \item MMAN \cite{ma2019deep}, which introduces semantic multi-modality representation learning into adversarial domain adaptation and captures fine-grained category information by multi-channel constraint. 
  \item CDAN \cite{long2018conditional}, which aligns the conditional distribution in adversarial learning.
  \item DSBN \cite{chang2019domain}, which adopts the domain-specific batch normalization in models.
  \item BSP \cite{chen2019transferability}, which improves the feature discriminability by penalizing the largest singular values.
  \item BNM \cite{cui2020towards}, which achieves the discriminability and diversity of the predictions with batch nuclear-norm maximization. 
  \item MDD \cite{zhang2019bridging}, which proposes a hypothesis-induced discrepancy for domain adaptation.
  \item GVB-GD \cite{cui2020gradually}, which proposes a gradually vanishing bridge mechanism for adversarial-based domain adaptation.
  \item GSDA \cite{hu2020unsupervised}, which aims to learn domain invariant representations by hierarchical domain alignment.
  \item STAR \cite{lu2020stochastic}, which tries to employ more classifiers by sampling from Gaussian distribution without more parameters.
  \item CAN \cite{kang2019contrastive}, which introduces class information into domain alignment by minimizing the contrastive domain discrepancy.
  \item SHOT \cite{liang2020we}, which develops a framework for data-free UDA based on hypothesis transfer learning.
  \item ModelAdapt \cite{li2020model}, which adopts a collaborative class conditional generative adversarial networks to avoid using source data.
  \end{enumerate*}

Among these methods, SHOT and ModelAdapt are developed under the source data-free UDA setting. We implement our method under both standard UDA and source data-free UDA settings.

\subsection{Implementation Details}
\label{sec:imple}

\textbf{Network architecture}. We adopt a ResNet-50 (for Office-31) and ResNet-101 (for VisDA-2017) pre-trained on ImageNet \cite{deng2009imagenet} as the feature encoder in the experiments of the standard UDA task. We replace the last FC layer with the task-specific FC classifier layer. All the network parameters are shared between different domains except those of the batch normalization (BN) layers as we utilize the domain-specific BN \cite{chang2019domain}. Under the source data-free UDA setting, following \cite{liang2020we}, we use one bottleneck layer containing a FC layer and a BN layer after convolutional layers in the feature encoder module $g$. Furthermore, we remove the bias of the task-specific FC classifier layer and perform normalization for the classifier.

\textbf{Training details}. The network is trained by using mini-batch SGD with a momentum of 0.9. The initial learning rate $\eta_0$ is set as $1e^-3$ for pre-trained convolutional layers and $1e^-2$ for newly added layers. We employ the same learning rate scheduler $\eta = \eta_0 \cdot (1 + 10 \cdot p)^{-b}$ as  \cite{long2015learning,long2017deep,ganin2015unsupervised}, where $p$ denotes training process linearly increase from 0 to 1. Following  \cite{kang2019contrastive}, for Office-31, $b=0.75$ while for VisDA-2017, $b=2.25$. When pre-training models with source samples under source data-free setting, following  \cite{liang2020we}, we randomly split the source dataset into 0.9/0.1 train-validation sets and select the optimal source pre-trained model based on the validation set. We use one RTX 3090 with 24GB for experiments.

\begin{table}[t]
  \caption{Accuracy(\%) on Office-31 for unsupervised domain adaptation (ResNet-50). $\dagger$ denotes that this method is developed under the source data-free UDA setting.}
  \label{tab:office}
  \centering
\setlength{\tabcolsep}{0pt} 
   \begin{tabular*}{\linewidth}{@{\extracolsep{\fill}\quad}lccccccc}
    \toprule
    Method & A$\rightarrow$D & A$\rightarrow$W & D$\rightarrow$A & D$\rightarrow$W & W$\rightarrow$A & W$\rightarrow$D & Avg \\
    \midrule
    ResNet-50 \cite{he2016deep}    & 68.9 & 68.4 & 62.5 & 96.7 & 60.7 & 99.3 & 76.1 \\
    \midrule
    DAN \cite{long2015learning}            & 78.6 & 80.5 &  63.6 & 97.1 & 62.8 & 99.6 & 80.4 \\
    DANN \cite{ganin2015unsupervised}      &  79.7 &  82.0 &  68.2 & 96.9 & 67.4 & 99.1 & 82.2 \\
    JAN \cite{long2017deep}     &  84.7 &  85.4 & 68.6 & 97.4 & 70.0 & 99.8 &  84.3 \\
    MMAN \cite{ma2019deep}      &  85.8 &  85.8 & 70.3 & 97.4 & 71.2 & \textbf{100.} &  85.1 \\
    SAFN+ENT \cite{xu2019larger}     & 92.1 & 90.3 & 73.4 & 98.7 & 71.2 & \textbf{100.} & 87.6 \\
    CDAN \cite{long2018conditional}      & 92.9 & 94.1 & 71.0 & 98.6 & 69.3 & \textbf{100.} & 87.7 \\
    CDAN+BSP \cite{chen2019transferability}    & 93.0 & 93.3 & 73.6 & 98.2 & 72.6 & \textbf{100.} & 88.5 \\
    CDAN+BNM \cite{cui2020towards}      & 92.9 & 92.8 & 73.5 &  98.8 & 73.8 & \textbf{100.} & 88.6 \\
    MDD \cite{zhang2019bridging}           & 93.5 & 94.5 & 74.6 &  98.4 & 72.2 & \textbf{100.} & 88.9 \\
    GVB-GD \cite{cui2020gradually}         & 95.0 & 94.8 & 73.4 & 98.7 &  73.7 & \textbf{100.} & 89.3 \\
    GSDA \cite{hu2020unsupervised}           & 94.8 & 95.7 & 73.5 & 99.1 &  74.9 & \textbf{100.} & 89.7 \\
    CAN \cite{kang2019contrastive}         & 95.0 & 94.5 &  \textbf{78.0} & 99.1 & \textbf{77.0} & 99.8 & \textbf{90.6} \\
    \midrule
    CDCL (ours)     & \textbf{96.0} & \textbf{96.0} & 77.2 & \textbf{99.2} & 75.5 & \textbf{100.} & \textbf{90.6} \\
    \midrule
    SHOT \cite{liang2020we} $\dagger$        & 94.0 & 90.1 &  74.7 & 98.4 & 74.3 & 99.9 & 88.6  \\
    ModelAdapt \cite{li2020model} $\dagger$     & 92.7 & \textbf{93.7} & 75.3 & \textbf{98.5} & \textbf{77.8} & 99.8 & \textbf{89.6}  \\
    CDCL (ours) $\dagger$     & \textbf{94.4} & 92.1 &  \textbf{76.4} & \textbf{98.5} & 74.1 & \textbf{100} & 89.3  \\
    \bottomrule
  \end{tabular*}
\end{table}

\begin{figure*}[t]%
    \centering
    \subfloat[\centering Features without adaptation]{{\includegraphics[width=0.44\linewidth]{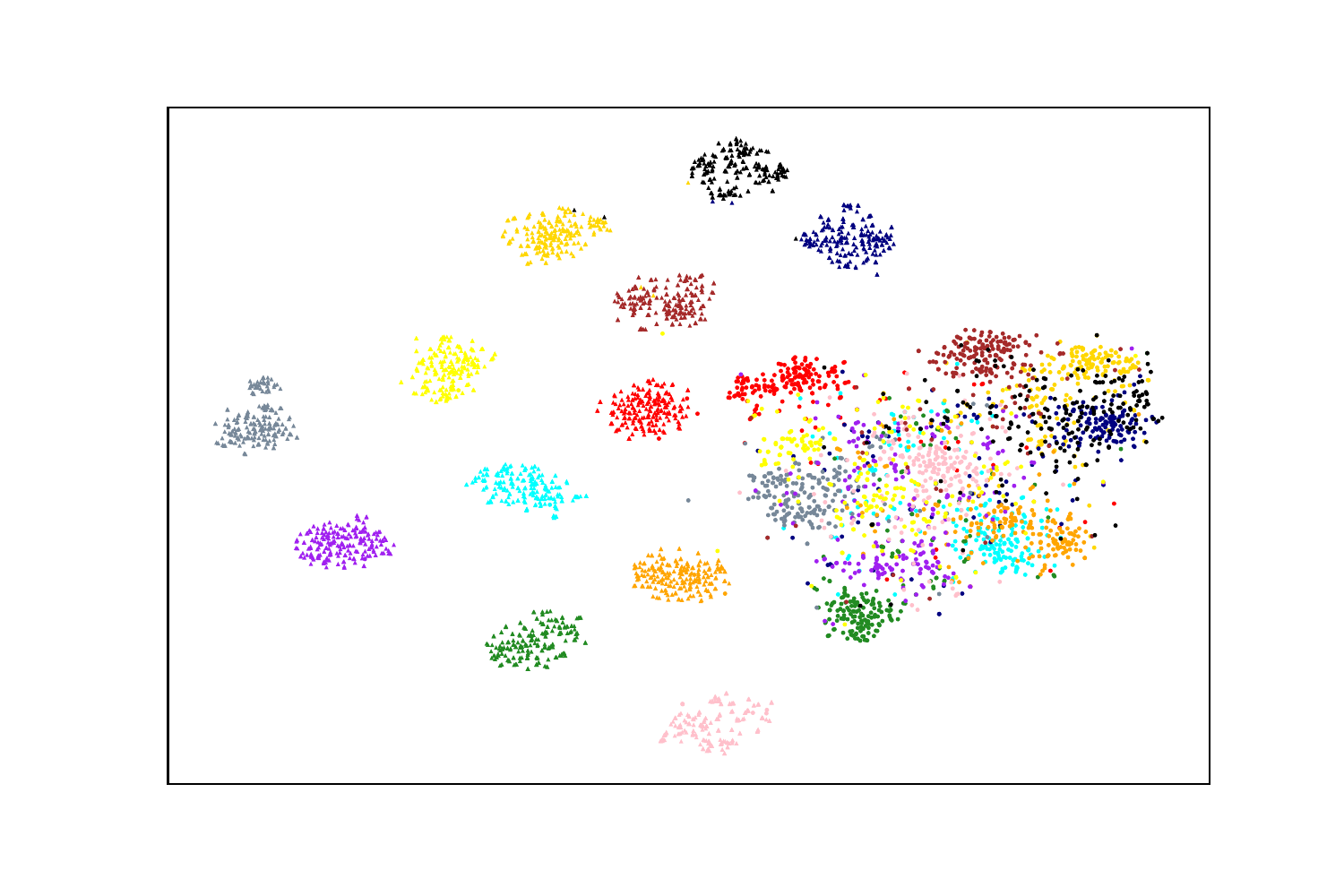} }}%
    \qquad
    \subfloat[\centering  Features aligned by \system.]{{\includegraphics[width=0.44\linewidth]{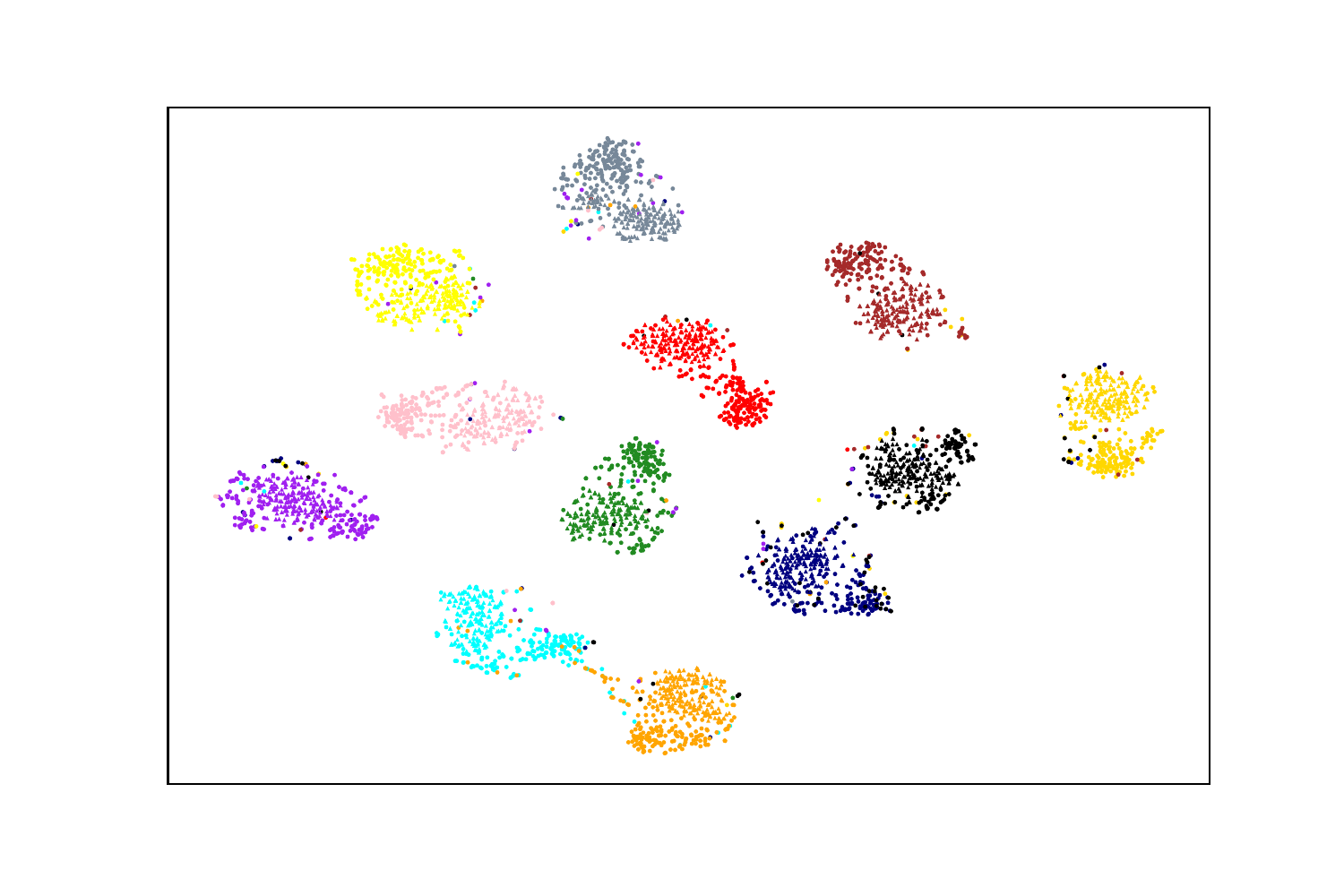} }}%
    \vspace{-0.08in}
    \caption{The t-SNE \cite{van2008visualizing} visualization of features from the source domain and the target domain before and after alignment. The triangle and circle markers indicate the source and target samples respectively, and different colors denote different classes.}%
    \label{fig:example}%
\end{figure*}

\begin{figure*}[t] \centering
  \resizebox{\linewidth}{!}{\includegraphics[width=0.92\linewidth]{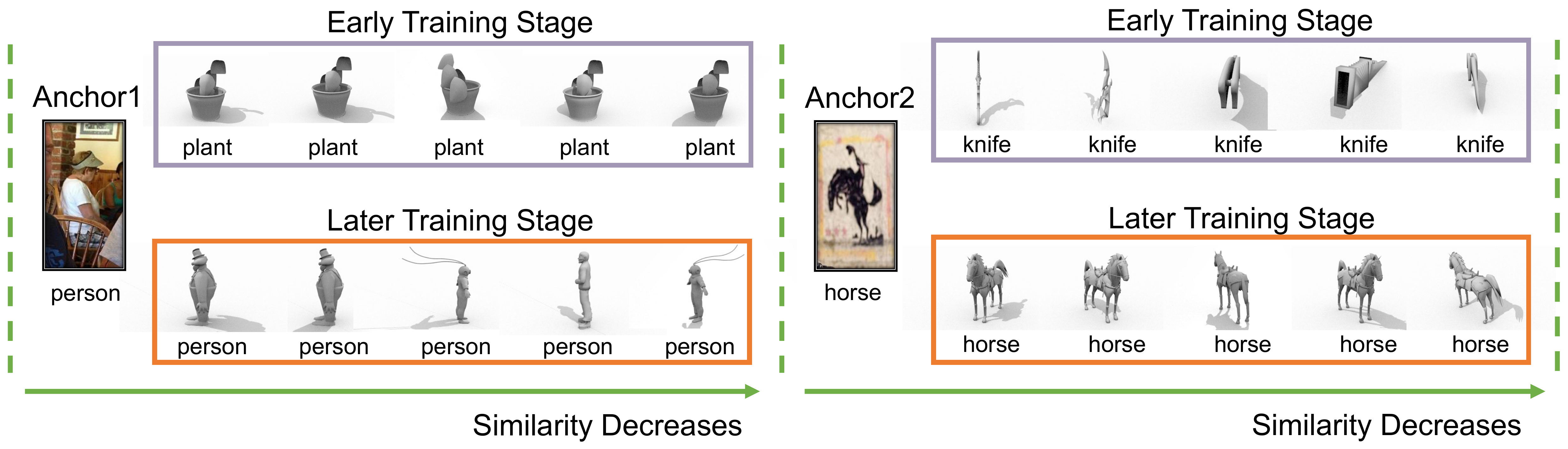}}
      \vspace{-0.08in}

  \caption{Given an anchor from one domain, top-5 cross-domain samples are retrieved by comparing feature similarity after the first epoch of training (Top) and at the end of training (Bottom).}
  \label{fig:knn}
\end{figure*}

\subsection{Main Results}
Table~\ref{tab:visda} and Table~\ref{tab:office} summarize the results of our approach and comparisons with state-of-the-art methods. On both datasets, We can see that all domain adaptation methods achieve significantly better results compared to the source-only method, confirming the importance of feature alignment. On VisDA, we observe \system achieves a mean accuracy of 88.6\% across all categories, outperforming all state-of-the-art approaches and boosting the accuracy of source-only baseline by 26.2\%. In particular, \system is better than CAN \cite{kang2019contrastive} by 1.4\% absolute point and beats JCL \cite{park2020joint} by 1.0\%, which is notable given that VisDA is a challenging benchmark.

When the source data are no longer available, \system achieves a mean accuracy of 87.5\%, outperforming the state-of-the-art approach SHOT \cite{liang2020we} by 4.6\% point, highlighting the effectiveness of our approach. Comparing the data-free setting and the conventional UDA setting, we see that \system is slightly (0.9\%) worse in the data-free setting. It is noteworthy that \system for source data-free setting still surpasses many UDA methods for standard setting. This suggests that one can simply explore a model trained on the source domain for effective transfer without the need to access the source data.

We observe similar trends on Office-31 under both settings. In particular, in the conventional UDA setting, \system offers a mean accuracy of 90.6\% across 6 different tasks, which is on par with the best results in the literature. Under the data-free setting, \system achieves a mean accuracy of 89.3\%, comparable to ModelAdapt. It is worth mentioning that results are better on Office-31 compared to VisDA since the dataset is smaller. Besides, due to the small domain gap between D and W, it is easy to achieve high accuracy on Office-31 tasks D$\rightarrow$W and W$\rightarrow$D, even for source-only models. Since domains \textbf{W} and \textbf{D} contain fewer samples compared to domain \textbf{A}, the performance on tasks W$\rightarrow$A and D$\rightarrow$A is relatively poor.

\begin{figure*}[t]%
    \centering
    \subfloat[\centering accuracy on VisDA-2017]{{\includegraphics[width=0.4\linewidth]{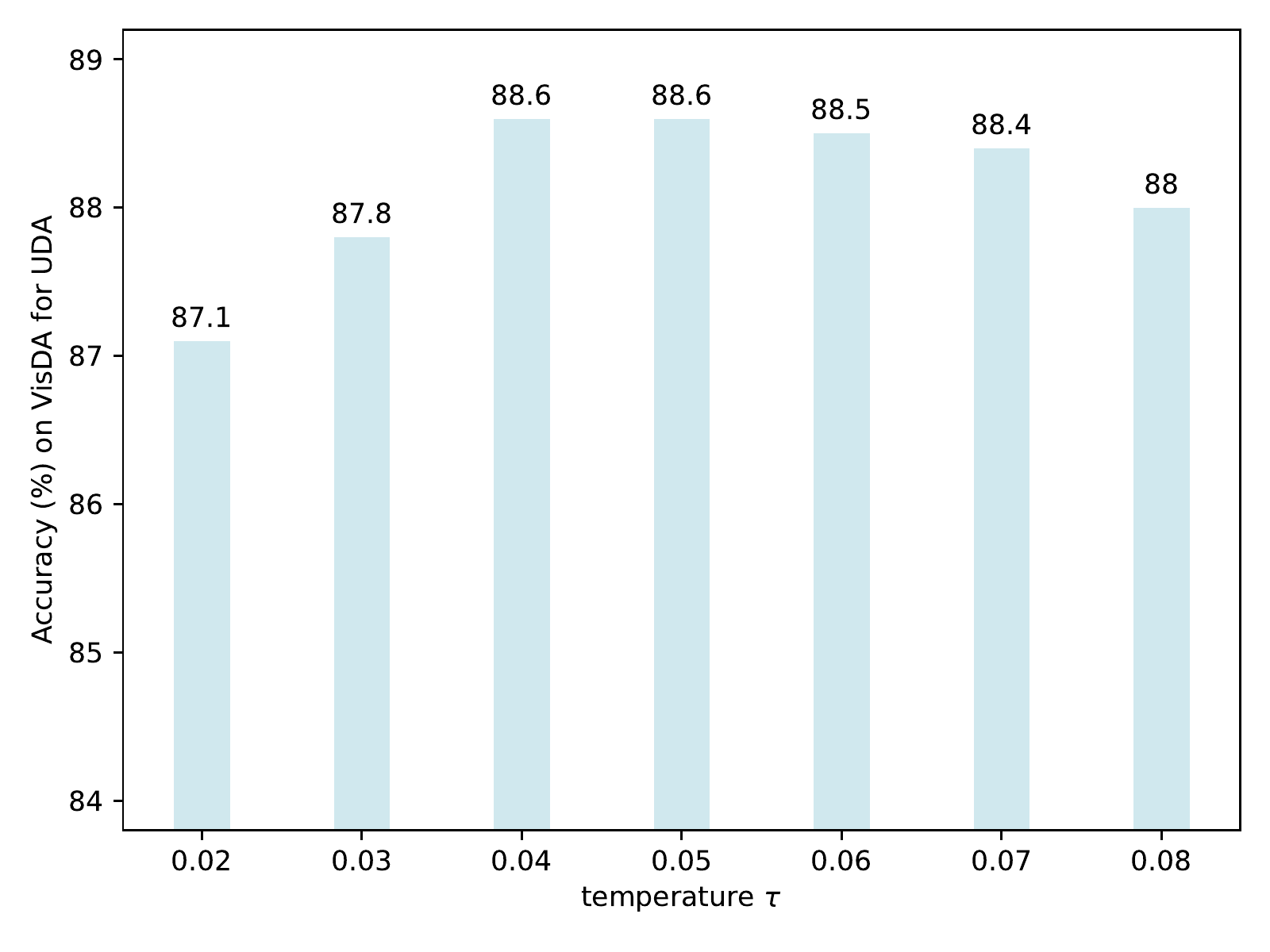} }}%
    \qquad
    \subfloat[\centering  accuracy on VisDA-2017]{{\includegraphics[width=0.4\linewidth]{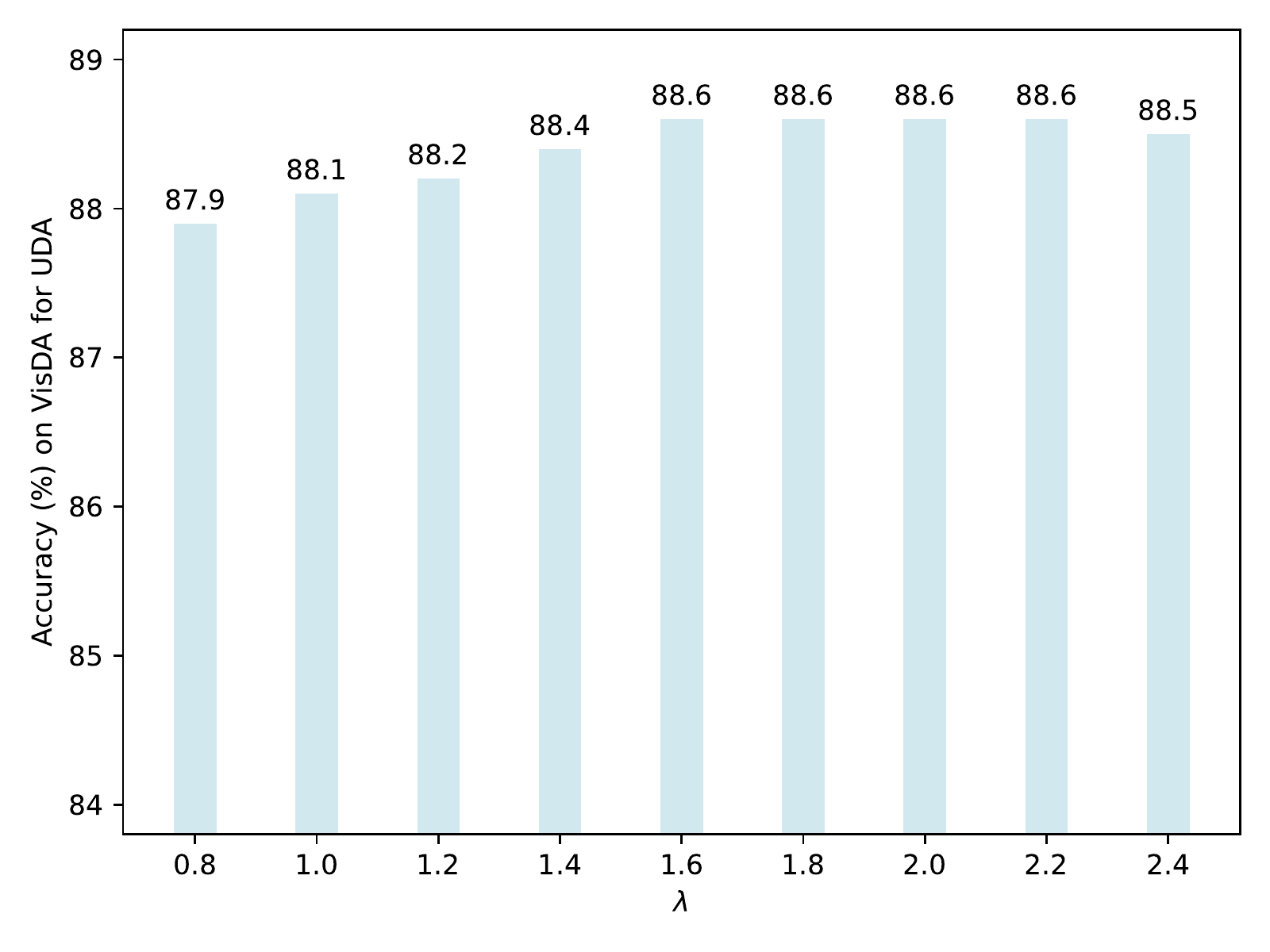} }}%
    \caption{Performance sensitivity of 2 hyper-parameters $\tau$, $\lambda$ in CDCL. }%
    \label{fig:hyperparameters}%
\end{figure*}

\subsection{Ablation Studies and Discussions}
In this section, we conduct a set of ablation experiments to justify the effectiveness of different components and provide discussions.

\textbf{Positive and Negative Pairs.} We mainly form positive pairs and negative pairs using cross-domain samples. We now discuss alternative ways to form pairs and report results on VisDA. In particular, we use the following approach to form pairs: (1) In-domain, where anchors come from both domains but samples that are used to form pairs are from the same domain and same category as the anchor; (2) Combined-domain, where the source and target domains are mixed and pairs are generated by simply considering label information; (3) Cross-domain ($\mathcal{L}^{s,i}_{CDC}$ only), which simply $\mathcal{L}^{s,i}_{CDC}$ in Eqn.~\ref{eqn:cdc_bi} using samples in the source domain as anchors; (4) Cross-domain ($\mathcal{L}^{t,i}_{CDC}$ only), which uses samples in the target domain as anchors. From the results shown in Table~\ref{tab:ablation-loss}, we observe that cross-domain alignment achieves better results compared to performing a simply in-domain alignment. This suggests the importance of forming pairs from two domains in order to produce domain-invariant features. In addition, we observe that mixing both domains together is worse than \system, possibly due to the fact that jointly modeling intra-class and inter-class information is challenging. Moreover, the bi-directional use of anchors is better compared to using anchors simply from one domain.

\begin{table}
  \caption{Ablation study for the selection of anchors, positive samples and negative samples in contrastive loss.}
  \label{tab:ablation-loss}
  \centering
\setlength{\tabcolsep}{0pt} 
   \begin{tabular*}{\linewidth}{@{\extracolsep{\fill}\quad}lcccc}
    \toprule
    Method & Anchor & Positive & Negative & VisDA \\
    \midrule
    In-domain       & all & same & same & 86.5 \\
    Combined-domain  & all & all & all & 87.3 \\
    Cross-domain ($\mathcal{L}^{s,i}_{CDC}$ only)    & source & different & different & 87.5 \\ 
    Cross-domain ($\mathcal{L}^{t,i}_{CDC}$ only)    & target & different & different & 86.6 \\
    \system     & all & different & different  & \textbf{88.6} \\
    \bottomrule
  \end{tabular*}
\end{table}

\textbf{The impact of hyper-parameters.} We test the sensitivity of CDCL to the temperature $\tau$ on VisDA-2017 for standard UDA. As shown in Figure~\ref{fig:hyperparameters}(a), the accuracies around $\tau = 0.05$ are not sensitive. When the $\tau$ grows larger, the accuracy steadily increases before decreasing. Additionally, we study the effect of $\lambda$ on VisDA. Results in Figure~\ref{fig:hyperparameters}(b) show that the accuracy around $\lambda = 1.6$ are also not sensitive and the gap of accuracies is smaller than 0.2 when $\lambda > 1.4$. Therefore, CDCL is not sensitive to its hyper-parameters.

\textbf{Feature Visualization.} We further use t-SNE \cite{van2008visualizing} to visualize features from the source and target domain before and after alignment in Figure~\ref{fig:example}. We can see that before alignment, source and target features are separated into two clusters, which demonstrates the gap between the two domains. After alignment with \system, we see that features from different domains are mixed together and features from different classes are well separated.

\textbf{Learned Feature Distance.} We visualize in Figure~\ref{fig:knn} top-5 retrieved images given an anchor image at the beginning and the end of training. We observe that in early states, retrieved samples are similar to the anchor in shape but belong to a different category. As training continues, \system gradually pulls features from the same class to be closer. This highlights the effectiveness of \system in learning domain-invariant features.

\section{Conclusion}
 In this paper, we presented \system, a simple yet effective framework for unsupervised domain adaptation. \system builds upon contrastive learning to align features for domain alignment and is suitable for both the standard UDA setting and the source data-free setting. In particular, given an image from one domain, we minimize its distance with respect to samples in the same class but from a different domain relative to all other cross-domain samples from different categories. Since labels are not available for the target domain, we generate pseudo labels using clustering. Further, we showed that \system can be easily adapted to the source data-free settings through considering classifier weights as class prototypes. We conducted extensive experiments on two widely used domain adaptation benchmarks and demonstrated that \system achieves state-of-the-art performance on both datasets.

\section*{Acknowledgment}

This project was supported by NSFC under Grant No. 62102092.

\ifCLASSOPTIONcaptionsoff
  \newpage
\fi


\bibliographystyle{IEEEtran}
\bibliography{IEEEfull}

%

\begin{IEEEbiography}[{\includegraphics[width=1in,height=1.25in,clip,keepaspectratio]{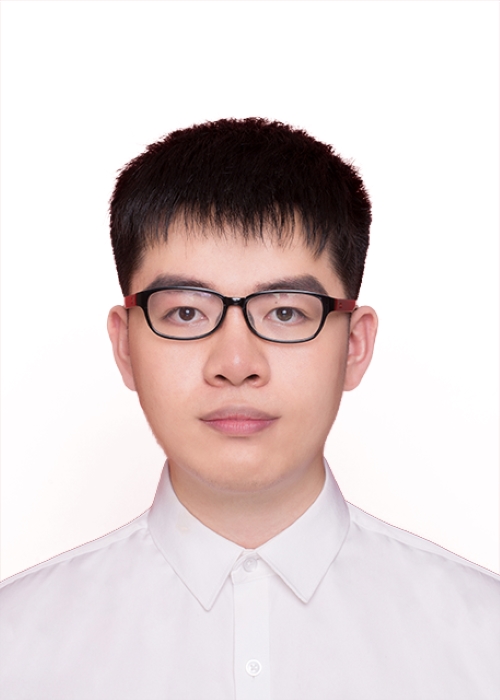}}]{Rui Wang} 
received the B.S. degree from Fudan University, Shanghai, China, in 2021. He is currently pursuing his Ph.D. degree in Computer Science at Fudan University. His research interests include video understanding, unsupervised learning and domain adaptation.
\end{IEEEbiography}

\begin{IEEEbiography}[{\includegraphics[width=1in,height=1.25in,clip,keepaspectratio]{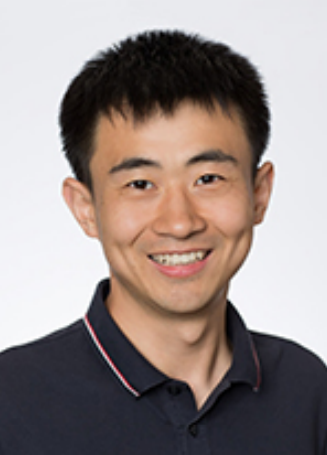}}]{Zuxuan Wu} 
received his Ph.D. in Computer Science from the University of Maryland with Prof. Larry Davis in 2020. He is currently an Associate Professor in the School of Computer Science at Fudan University. His research interests are in computer vision and deep learning. His work has been recognized by an AI 2000 Most Influential Scholars Honorable Mention in 2021, a Microsoft Research PhD Fellowship in 2019 and a Snap PhD Fellowship in 2017.
\end{IEEEbiography}

\begin{IEEEbiography}[{\includegraphics[width=1in,height=1.25in,clip,keepaspectratio]{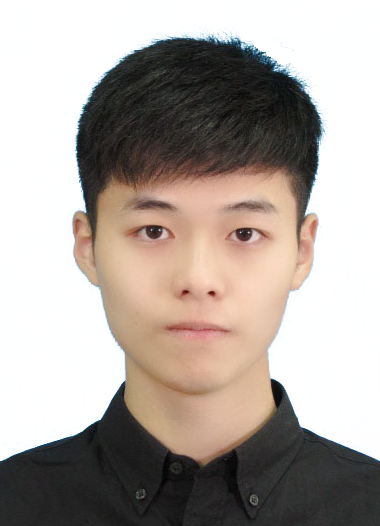}}]{Zejia Weng} 
received the B.S. degree from Fudan University, Shanghai, China in 2020. He is currently pursuing his M.S. degree in the School of Computer Science at Fudan University. His research interests are in computer vision and deep learning, especially large-scale video understanding.
\end{IEEEbiography}

\begin{IEEEbiography}[{\includegraphics[width=1in,height=1.25in,clip,keepaspectratio]{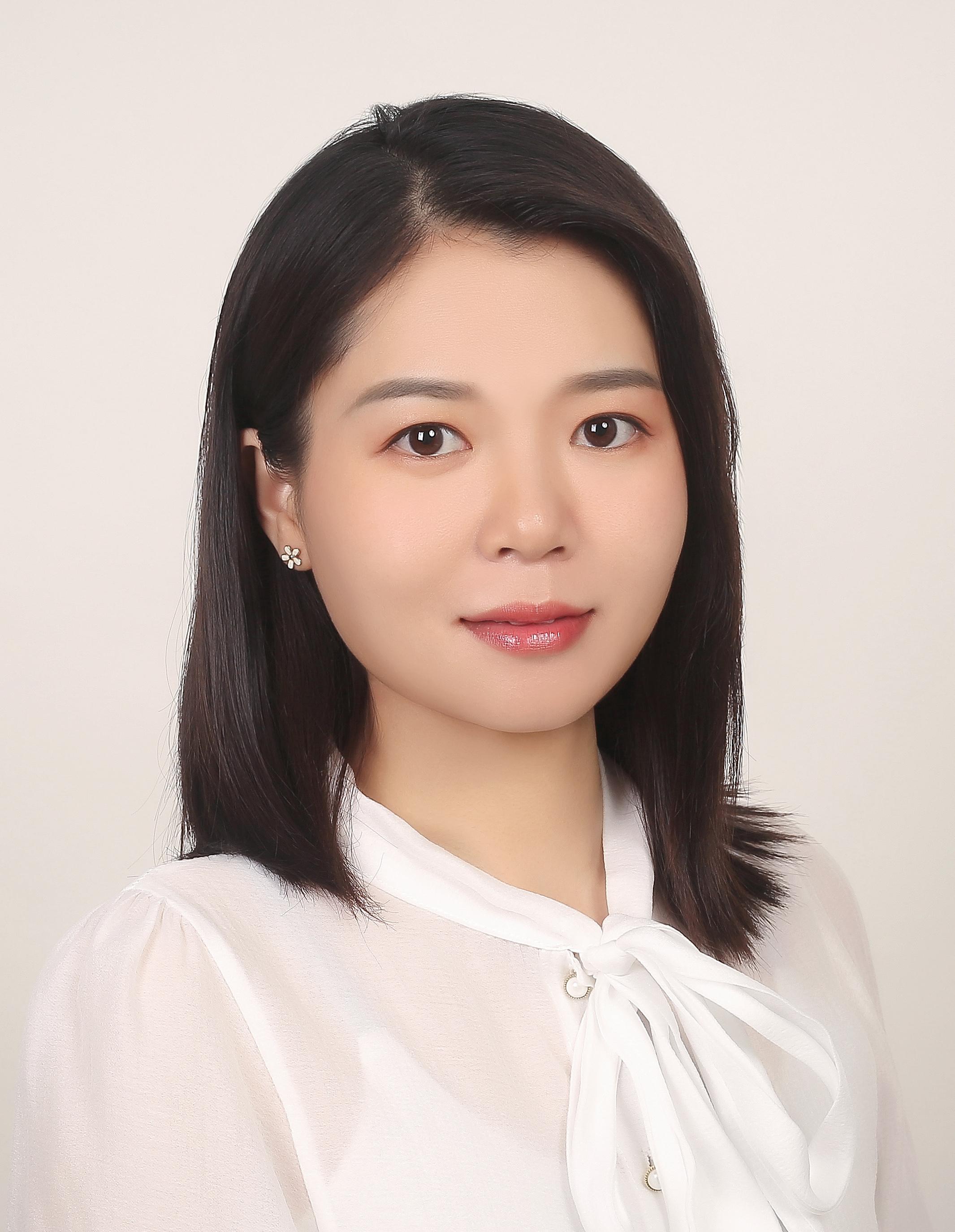}}]{Jingjing Chen} 
is now a pre-tenured associate professor at the School of Computer Science, Fudan University. Before joining Fudan University, she was a postdoc research fellow at the School of Computing in the National University of Singapore. She received her Ph.D. degree in Computer Science from the City University of Hong Kong in 2018. Her research interest lies in diet tracking and nutrition estimation based on multi-modal processing of food images, including food recognition, cross-modal recipe retrieval.
\end{IEEEbiography}

\begin{IEEEbiography}[{\includegraphics[width=1in,height=1.25in,clip,keepaspectratio]{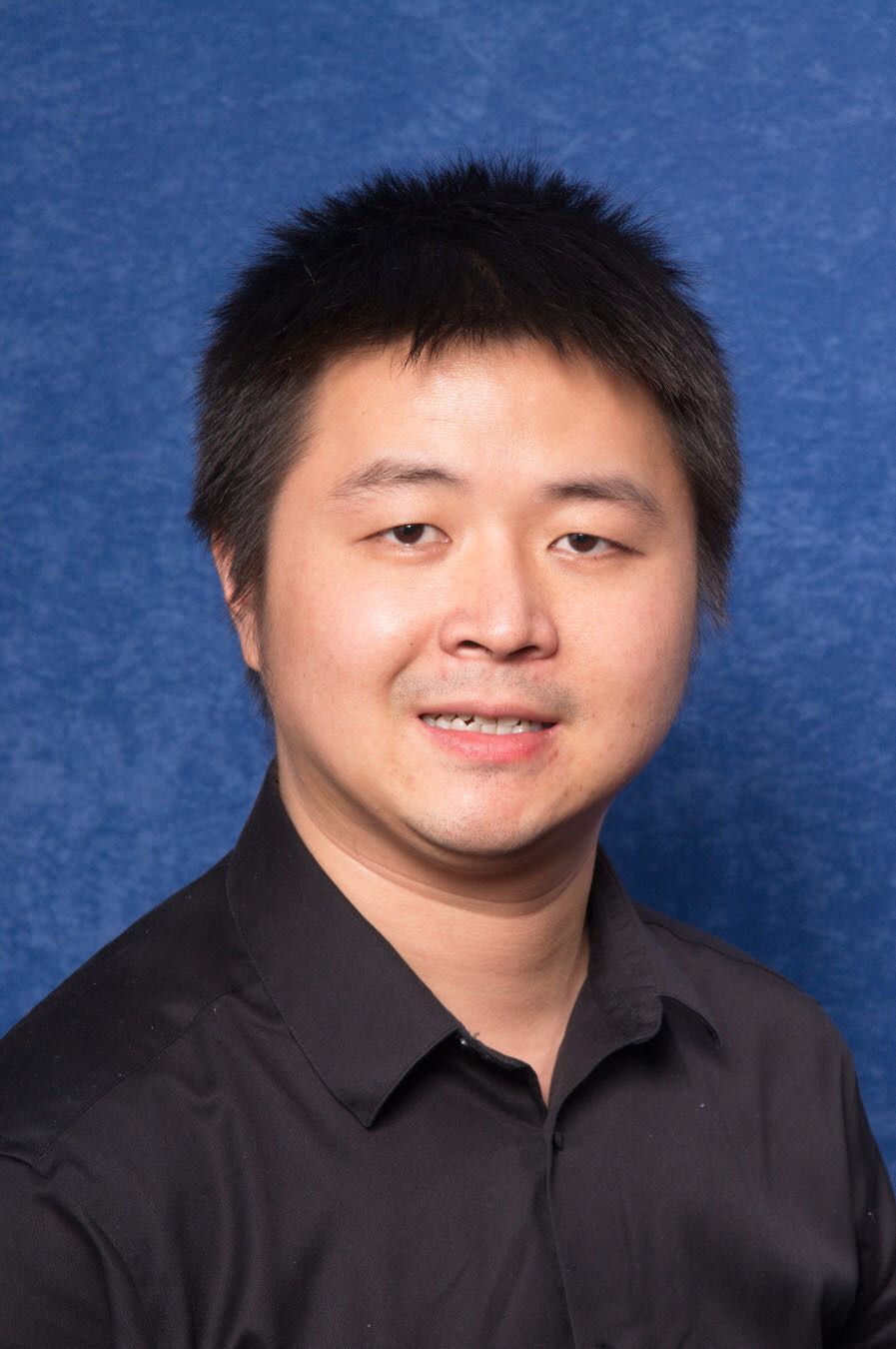}}]{Guo-Jun Qi}
 (M14-SM18-F22)  is the Chief Scientist and the director of Seattle Research Center in the OPPO Research USA since 2021. Before that, he is the Chief Scientist who led and oversaw an international R\&D team in the domain of multiple intelligent cloud services, including smart cities, visual computing service, medical intelligent service, and connected vehicle service at Futurewei since 2018. He was a faculty member in the Department of Computer Science and the director of MAchine Perception and LEarning (MAPLE) Lab at the University of Central Florida since 2014. Prior to that, he was also a Research Staff Member at IBM T.J. Watson Research Center, Yorktown Heights, NY.
\end{IEEEbiography}

\begin{IEEEbiography}[{\includegraphics[width=1in,height=1.25in,clip,keepaspectratio]{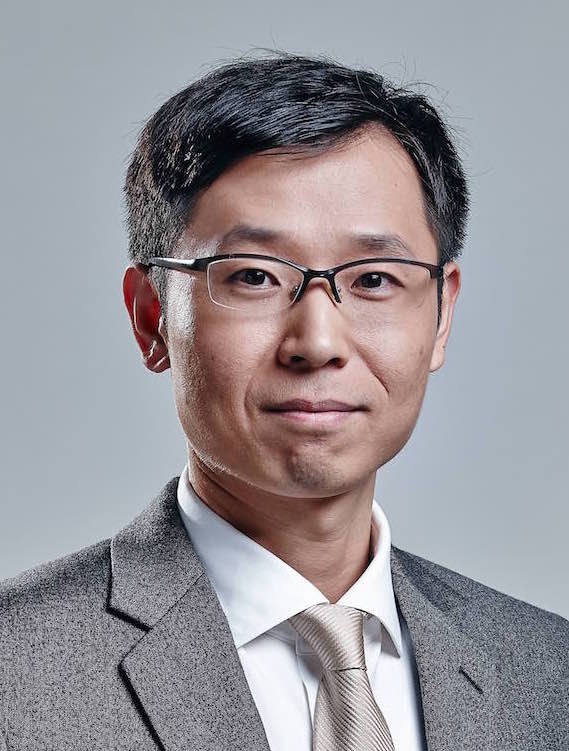}}]{Yu-Gang Jiang} 
received the Ph.D. degree in Computer Science from City University of Hong Kong in 2009 and worked as a Postdoctoral Research Scientist at Columbia University, New York during 2009-2011. He is currently a Professor and Dean at School of Computer Science, Fudan University, Shanghai, China. His research lies in the areas of multimedia, computer vision and trustworthy AI. His work has led to many awards, including the inaugural ACM China Rising Star Award, the 2015 ACM SIGMM Rising Star Award, and the research award for excellent young scholars from NSF China. 
\end{IEEEbiography}

\end{document}